\documentclass[12pt]{article}

\usepackage{amsmath}
\usepackage{color}
\usepackage{amsfonts}
\usepackage{amssymb}

\usepackage{fullpage}
\usepackage{graphicx}
\usepackage{booktabs}

\usepackage[colorlinks,citecolor=blue]{hyperref}

\usepackage{amsthm}

\usepackage{algorithm}
\usepackage{algpseudocode}

\usepackage{subcaption}

\newtheorem{definition}{Definition}
\newtheorem{proposition}{Proposition}

\date{}

\begin{document}

\title{An Aggregation Method for Sparse Logistic Regression}

\author{Zhe Liu\thanks{Corresponding author. Email: zheliu@galton.uchicago.edu. Mailing address: University of Chicago, Department of Statistics, 5734 S. University Avenue, Chicago, IL 60637}}

\maketitle
\begin{abstract}
$L_1$ regularized logistic regression has now become a workhorse of data mining and bioinformatics: it is widely used for many classification problems, particularly ones with many features. However, $L_1$ regularization typically selects too many features and that so-called false positives are unavoidable. In this paper, we demonstrate and analyze an aggregation method for sparse logistic regression in high dimensions. This approach linearly combines the estimators from a suitable set of  logistic models with different underlying sparsity patterns and can balance the predictive ability and model interpretability. Numerical performance of our proposed aggregation method is then investigated using simulation studies. We also analyze a published genome-wide case-control dataset to further evaluate the usefulness of the aggregation method in multilocus association mapping.
\end{abstract}


\section{Introduction}\label{sec:intro}

Logistic regression (LR) is now a widely used classification method in many fields such as machine learning, data mining, bioinformatics (Shevade and Keerthi, 2003; Liao and Chin, 2007; Wu et~al., 2009). In particular, the recent progress in multilocus association mapping of disease genes has been propelled by logistic regression using cases and controls. Hundreds of thousands of single nucleotide polymorphisms (SNPs) are being typed on thousands of individuals. When there are a large number of features to be learned, logistic regression is prone to over-fitting. It is well known that $L_1$ regularized logistic regression is a promising approach to reduce over-fitting, and can be used for feature selection in the presence of many irrelevant features  (Ng, 2004; Goodman, 2004; Lee et~al., 2006). The $L_1$ penalty is an effective device for variable selection, especially in problems where the number of features $p$ far exceeds the number of observations $n$. However, $L_1$ regularization typically selects too many variables and that so-called false positives are unavoidable (van de Geer et~al., 2011). That is, there might be too many irrelevant features are present in the selected models. \\

Given a collected family of estimators, linear or convex aggregation methods are another class of techniques to address model selection problems and provide flexible ways to combine various models into a single estimator (Rigollet and Tsybakov, 2011).  The idea of aggregation of estimators was originally described in Nemirovski (2000). The suggestion put forward by Nemirovski (2000) is to achieve two independent subsamples from the original sample by randomization; individual estimators are constructed from the first subsample while the second is used to perform aggregation on those individual estimators (Rigollet and Tsybakov, 2012). This idea of two-step procedures carries over to models with i.i.d. observations where one can do sample splitting. Along with this method, one might consider aggregate estimators using the same observations for both estimation and aggregation. However, this would generally result in overfitting. Also, notice that such direct sample splitting does not apply to independent samples that are not identically distributed as in the present setup. \\

A primary motivation for aggregating estimators is that it can improve the estimation risk, as ``betting'' on multiple models can provide a type of insurance against a single model being poor (Leung and Barron, 2006). Most of the recent work on estimator aggregation deals with regression learning problems. For example, exponential screening for linear models provides a form of frequentist averaging over a large model class, which enjoys strong theoretical properties (Rigollet and Tsybakov, 2011). An aggregation classifier is proposed in Lecu\'e (2007) and an optimal rate of convex aggregation for the hinge risk is also obtained. \\

In this paper, we propose a novel estimating procedure for the regression coefficients in logistic models by considering a linear combination of various estimators with different underlying sparsity patterns. A sparsity pattern is defined as a binary vector with each element indicating whether the corresponding feature is absent or not. Given any sparsity pattern, we would use a single logistic regression to obtain the associated individual estimator. The corresponding component weights for individual estimators are determined to ensure a bounded risk of the aggregation estimator. \\

Our aggregation procedure is based on the sample-splitting: after partitioning the initial sample into two subsamples by randomization, the first subsample is set to construct the estimators and the second subsample is then used to determine the weights and aggregate these estimators. To carry out the analysis of the aggregation step, it is enough to work conditionally on the first subsample so that the problem reduces to aggregation of deterministic functions (Rigollet and Tsybakov, 2012). Namely, given a set of deterministic estimator $\theta_m$ with sparsity pattern $m$, one can construct aggregation estimator $\hat{\theta}$ satisfying the following oracle inequalities
\begin{align}
\mathbb{E}R(\hat{\theta})\leq C \min_{m\in\mathcal{M}}R(\theta_m)+\delta_{n,p},
\end{align}
where $R(\cdot)$ is an empirical risk function, $\mathbb{E}$ denotes the expectation, $C\geq 1$ is a constant, $\mathcal{M}$ is a family of sparsity patterns, and $\delta_{n,p}\geq 0$ is a small remainder term characterizing the performance of aggregation.  Ideally, we wish to find an aggregation estimator whose risk is as close as possible (in a probabilistic sense) to the minimum risk of the individual estimators. \\

The rest of the paper is organized as follows. In Section 2, we describe the logistic aggregation estimator in detail. Section 3 evaluates the method on simulated data. Section 4 applies the method to real data on genome-wide case-control association study. We provide further discussion in Section 5.

\section{Methods}

In this section, we consider data pairs $\mathcal{D}_n:=\{(x_i, y_i)\}_{i=1}^n$ of covariates and responses, sampled from a population, where each $x_i$ is a $p$-dimensional vector, and $y_i\in\{0,1\}$ is a class label. In case-control studies, the dichotomous response variable $y_i$ is typically coded as 1 for cases and 0 for controls. Assume these pairs satisfy a general setting that 
\begin{align}
\eta(x_i):&=\text{Pr}(y_i=1|x_i;\theta) \\
&=\sigma(x_i^T\theta)=\frac{1}{1+\exp(-x_i^T\theta)},
\end{align}
where $\theta\in\mathbb{R}^p$ is a vector of parameters for regression coefficients, and $\sigma(\cdot)$ is the sigmoid function. \\

For logistic regression, the parameter vector $\theta$ is usually estimated by maximizing the log-likelihood
\begin{align}
l(\theta; \mathcal{D}_n):=\sum_{i=1}^{n} \Bigl[y_i\theta^Tx_i-\log(1+\exp(\theta^Tx_i))\Bigr].
\end{align}

To encourage sparse solutions, we can subtract a lasso penalty $g(\theta)$ from the log-likelihood
\begin{align}
g(\theta):=\lambda\sum_{j=1}^p|\theta_j|,
\end{align}
where $\lambda$ is a tuning parameter. For a given value of the tuning constant $\lambda$, maximizing the penalized log-likelihood singles out a certain number of predictors with non-zero regression coefficients. Wu et~al. (2009) proposed an efficient algorithm called cyclic coordinate ascent in maximizing the lasso penalized log-likelihood.

\subsection{Logistic Aggregation Estimator}

Instead of picking a single final model, a more general approach to account for model uncertainty is to combine many models together, resulting in an ensemble model. We propose a new estimating procedure for the regression coefficients in logistic models by considering a linear combination of various estimators with different underlying sparsity patterns. We call a sparsity pattern any binary vector $m\in\{0,1\}^p$. The $i$th coordinate of $m$ can be interpreted as indicators of presence ($m_i=1$) or absence ($m_i=0$) of the $i$th feature. We denote by $|m|_1$ the number of ones in $m$. \\

The aggregation is based on a random splitting of the sample into two independent subsamples $\mathcal{D}_{n_1}^{(1)}$ and $\mathcal{D}_{n_2}^{(2)}$ of size $n_1$ and $n_2$, respectively, where $n_1+n_2=n$. The first subsample $\mathcal{D}_{n_1}^{(1)}$ is used to construct individual estimators and the second subsample $\mathcal{D}_{n_2}^{(2)}$ is then used to aggregate them. In what follows we will denote
\begin{align}
\mathcal{D}_{n_1}^{(1)}:&=\{(x_i^{(1)},y_i^{(1)})\}_{i=1,\ldots,n_1}, \\
\mathcal{D}_{n_2}^{(2)}:&=\{(x_i^{(2)},y_i^{(2)})\}_{i=1,\ldots,n_2}.
\end{align}

For each sparsity pattern $m$, we consider a logistic regression estimator of the true regression coefficient $\theta$ with constraints on the sparsity pattern represented by $m$
\begin{align}
\hat{\theta}_m:=\underset{\{\theta:\;\theta_i=0\text{ for any }m_i=0\}}{\text{argmax}}l(\theta; \mathcal{D}_{n_1}^{(1)}),
\end{align}
where the log-likelihood
\begin{align}
l(\theta; \mathcal{D}_{n_1}^{(1)}):=\sum_{i=1}^{n_1} \Bigl[y_i^{(1)}\theta^Tx_i^{(1)}-\log(1+\exp(\theta^Tx_i^{(1)}))\Bigr].
\end{align}

Note that each $\hat{\theta}_m$ is constructed using only data $\mathcal{D}_{n_1}^{(1)}$. The sparsity pattern $m$ determines the zero-patterns in the estimator $\hat{\theta}_m$ in the sense that the $i$th coordinate of $\hat{\theta}_m$ is zero if $m_i=0$. Working conditionally on the first subsample, we only need to consider the aggregation of deterministic estimators $\hat{\theta}_m$. \\

\begin{definition}
The logistic aggregation (LA) estimator is defined as a linear combination of the individual estimators $\hat{\theta}_m$'s
\begin{align}
\label{estimator}
\hat{\theta}^{\emph{\text{LA}}}:=\sum_{m\in\mathcal{M}}w_m\hat{\theta}_m,
\end{align}
where the data-determined weights are defined as
\begin{align}
\label{weight}
w_m:=\frac{\exp(l(\hat{\theta}_m; \mathcal{D}_{n_2}^{(2)}))\pi_m}{\sum_{m'\in\mathcal{M}}\exp(l(\hat{\theta}_{m'}; \mathcal{D}_{n_2}^{(2)}))\pi_{m'}}.
\end{align}
Here, $\pi_m$ is a (prior) probability distribution on the set of sparsity patterns $\mathcal{M}$, defined by
\begin{align}
\pi_m:=\frac{1}{H}\Bigl(\frac{|m|_1}{2ep}\Bigr)^{|m|_1},
\end{align}
where $e$ is the base of the natural logarithm, and $H$ is a normalization factor. \\
\end{definition}

From the definition, notice that the LA estimator linearly combines a set of logistic regression estimators with various underlying sparsity patterns. Based on this data-splitting technique, the first subsample is set to format these individual estimators $\hat\theta_m$'s for all $m\in\mathcal{M}$. Then the second subsample is used to determine the exponential weighting for each model or estimator. Note that the performance of each estimator is evaluated using an independent dataset, whose predictive likelihood would serve as its weight for model aggregation. \\

We also incorporate a deterministic factor $\pi_m$ into the weighted average to account for model complexity or model preference, in a manner that facilitates desirable risk properties (Leung and Barron, 2006). In this case, low-complexity models are favored. We require that the $\pi_m$'s and the $w_m$'s sum to one for the ease of Interpretation and theoretical concerns. \\

The exponential form of weight $w_m$ can be obtained as the solution of the following maximization problem. We define $\Lambda$ as a flat simplex
\begin{align}
\Lambda:=\Bigl\{\{\lambda_m\}_{m\in\mathcal{M}}:\;\lambda_m\geq 0,\;\sum_{m\in\mathcal{M}}\lambda_m=1\Bigr\}.
\end{align}
Then consider the following convex optimization problem
\begin{align}
\underset{\lambda\in\Lambda}{\text{argmax}}\Bigl\{\sum_{m\in\mathcal{M}}\lambda_m \cdot l(\hat{\theta}_m; \mathcal{D}_{n_2}^{(2)})-\text{KL}(\lambda;\pi)\Bigr\},
\end{align}
where $\pi:=\{\pi_m\}_{m\in\mathcal{M}}$ and $\text{KL}(\lambda;\pi)=\sum_{m\in\mathcal{M}}\lambda_m\log(\lambda_m/\pi_m)\geq0$. It follows from the Karush-Kuhn-Tucker (KKT) conditions that the $\{w_m\}_{m\in\mathcal{M}}$ defined in Definition~1 is exactly the unique solution to this convex optimization problem.

\subsection{Computational Implementation}

To implement the estimating procedure, note that exact computation of the LA estimator requires the calculation of at most $2^p$ individual estimators. In many cases this number could be extremely large, and we must make a numerical approximation. Observing that the LA estimator is actually the expectation of a random variable that has a probability mass proportional to $w_m$ on individual estimator $\hat\theta_m$ for $m\in\mathcal{M}$, the Metropolis-Hastings algorithm can be exploited to provide such an approximation. \\

\noindent \textbf{Algorithm 1.} \textit{The LA estimator can be approximated by running a Metropolis-Hastings algorithm on a $p$-dimensional hypercube:
\begin{itemize}
\item[(S1).] Initialize $u_t=\{0\}^p$, $t=0$; 
\item[(S2).] For each $t\geq0$, generate $u_t'$ following an uniform distribution on the neighbors of $u_t$ in the $p$-dimensional hypercube;
\item[(S3).] Generate an $[0,1]$-uniformly distributed number $r$;
\item[(S4).] Put $u_{t+1} \leftarrow u_t'$, if $r < \min\{1,\;w_{u_t'}/w_{u_t}\}$; otherwise, set $u_{t+1} \leftarrow u_t;$
\item[(S5).] Compute $\hat\theta_{u_{t+1}}$. Stop if $t>T_0+T$; 
otherwise, update $t \leftarrow t+1$ and go to step S2.
\end{itemize}
Then we can approximate $\hat\theta^{\emph{\text{LA}}}$ by 
\begin{align*}
\widehat{\hat{\theta}^{\emph{\text{LA}}}}=\frac{1}{T}\sum_{t=T_0+1}^{T_0+T}\hat\theta_{u_t},
\end{align*}
where $T_0\geq0$ and $T\geq0$ are arbitrary integers.} \\

The following proposition shows that the resulting Markov chain ensures the ergodicity. 
\begin{proposition}
The Markov chain $\{u_t\}_{t\geq 0}$ generated by Algorithm~1 satisfies
\begin{align}
\lim_{T\to \infty}\frac{1}{T}\sum_{t=T_0+1}^{T_0+T}\hat\theta_{u_t}=\sum_{m\in\mathcal{M}}w_m\hat{\theta}_m,\;\text{almost surely}.
\end{align}
\end{proposition}

The proof is straightforward as the Markov chain is clearly $w$-irreducible. The Metropolis-Hastings algorithm incorporates a trade-off between sparsity and prediction to decide whether to add or remove a feature. \\

Notice that the LA estimator itself would always give an estimate of $\theta$ in which all the elements are non-zero, since all the possible individual estimators are linearly mixed. However, the implementation of the Metropolis-Hastings algorithm would lead to a sparse estimate. Thus, such approximated aggregation estimator can also be used for the task of feature selection. \\

In high-dimensional settings where $p\gg n$, the Metropolis-Hastings algorithm takes the form of a random walk over the binary hypercube of the $2^p$ all-subsets models, which might not be computationally suitable. In such scenarios, it could be helpful to employ some pre-screening methods in order to approximate the model space and select a candidate set of features. Then our aggregation process is constructed on all-subsets models of this selected candidate set of features. For instance, $L_1$-penalized logistic regression could serve as a pre-screening tool to approximate the model space and select a candidate set of features. A similar technique is addressed in Fraley and Percival (2015), where Bayesian model averaging is combined with Markov chain Monte Carlo model composition by using the $L_1$ regularization path as the model space for approximation. Alternatively, the marginal effect of each feature can be tested one by one, and we exclude those unlikely to appear in the model. \\

We summarize the implementation of approximate algorithm for logistic aggregation procedure as follows. \\

\noindent \textbf{Algorithm 2.} 
\textit{Implementation of algorithm to approximate the LA estimator $\hat{\theta}^{\emph{\text{LA}}}$:
\begin{itemize}
\item[(S1).] Randomly partition the data into two sets $\mathcal{D}_{n_1}^{(1)}$ and $\mathcal{D}_{n_2}^{(2)}$;
\item[(S2).] Determine a suitable set of sparsity patterns represented by $\mathcal{M}$;
\item[(S3).] Apply Algorithm~1 to obtain the estimator $\widehat{\hat{\theta}^{\emph{\text{LA}}}}$.
\end{itemize}
}

\section{Simulation Studies}
In this section, we conduct simulation studies to evaluate the numerical performance of the proposed logistic aggregation (LA) estimator, competing with logistic regression (LR) with multiple testings and $L_1$-penalized logistic regression ($L_1$-LR). In addition, we include the elastic net method (Enet) (Zou and Hastie, 2005), where we subtract a elastic net penalty $h(\theta)$ from the log-likelihood
\begin{align}
h(\theta):=\frac{1-\alpha}{2}\cdot\lambda\sum_{j=1}^p\theta_j^2+\alpha\cdot\lambda\sum_{j=1}^p|\theta_j|,
\end{align}
where $0\leq\alpha\leq1$ is the elastic net mixing parameter. Note that $\alpha=1$ is the lasso penalty, and $\alpha=0$ the ridge penalty. \\

For the penalized logistic regression and elastic net methods, the tuning parameter $\lambda$ and elastic net mixing parameter $\alpha$ are chosen via cross-validations, as implemented in the R \texttt{glmnet} package (Friedman et~al., 2010). For the logistic regression method, each feature is tested using simple logistic regression one at a time, and then we use the Bonferroni correction to account for multiple testings at the significant level of 0.05. For the logistic aggregation estimator, half of dataset is randomly chosen for the construction of individual estimators and the other half is then used for aggregation. The set of sparsity patterns used in the Metropolis-Hastings algorithm is determined by applying the penalized logistic regression with cross-validations to the first subsample. \\

In the classification problem, we generate responses $y_i$ according to a logistic model
\begin{align}
y_i\sim\text{Bernoulli}\left(\frac{\exp(x_i^T\theta)}{1+\exp(x_i^T\theta)}\right).
\end{align}
Here, we fix a vector of true coefficients $\theta$ with only the first $p_0=5$ entries set to be nonzero, by putting $\theta=(2,2,2,2,2,0,0,\ldots,0)^T$. We consider the following scenarios for generating the covariate structures, using the R \texttt{huge} package (Zhao et~al., 2012)
\begin{itemize}
\item Independent model: the covariates $x_i$'s are generated by independent standard Gaussian distributions;
\item AR(1) model: the covariates $x_i$'s are generated by multivariate Gaussian distributions using the lag-1 autoregressive model as the inverse covariance matrix for the first 100 coordinates;
\item AR(2) model: the covariates $x_i$'s are generated by multivariate Gaussian distributions using the lag-2 autoregressive model as the inverse covariance matrix for the first 100 coordinates.
\end{itemize}

In each scenario, we draw $n=300$ data points on $p=5000$ or $10000$ features for training the estimators and another $3000$ independent data points for evaluating the out-of-sample prediction performance. \\

The goal is to compare the out-of-sample prediction and feature selection performance of those approaches in high-dimensional settings. The following evaluation criteria are considered in the comparisons:
\begin{itemize}
\item AUC: area under the receiver operating characteristic (ROC) curve, as a measure of out-of-sample classification performance;
\item FP: number of selected features that are actually false positive; the $i$th feature is considered to be selected by the estimator $\hat\theta$ if $|\hat\theta_i|>1/n;$
\item FN: number of selected features that are actually false negative;
\item Time: computational time in seconds. \\
\end{itemize}

We use 50 replications in such evaluations. For each criterion of evaluation, results are averaged over replications and the standard error is also reported. Table~\ref{sim} displays the results of simulation studies. We can see that the LA estimator has the highest scores of AUC in all scenarios. For the number of false negatives, notice that the LA, $L_1$-LR and Enet methods successfully identify almost all true features, while the LR method fails to identify nearly a half of true features under the AR models. In terms of the number of false positives, the $L_1$-LR and Enet estimator have much worse selection performance, where too many false positives are present; the LA estimator performs much better, although the LR method hardly results in any false positives. These methods do not differ much in the computational cost; the Enet method spends more time for cross-validations on two parameters. \\

\begin{figure}[!h]
\vspace{.2in}
\centering{\includegraphics[width=.675\textwidth,height=.3\textheight]{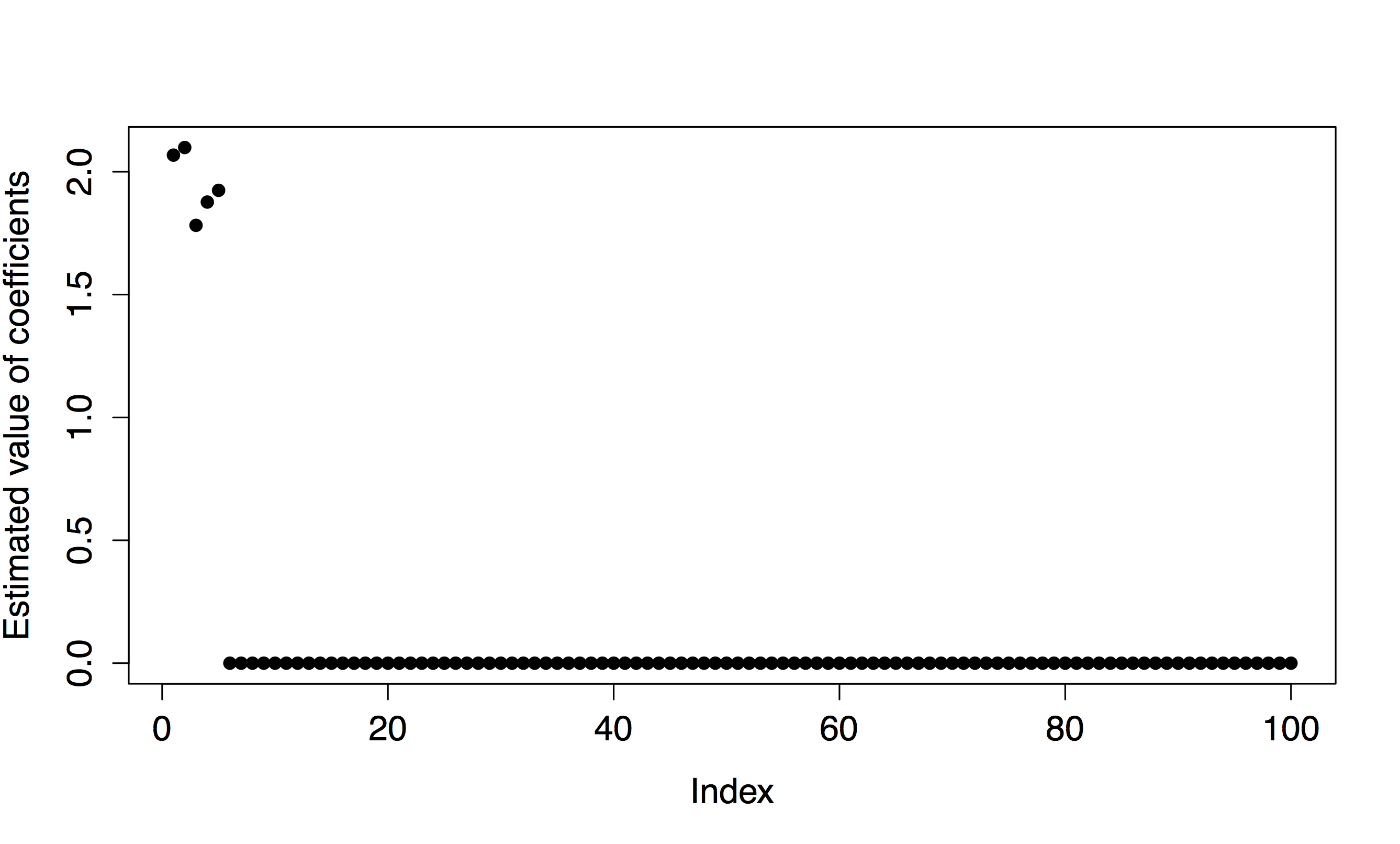}} \\
\centering{\includegraphics[width=.675\textwidth,height=.3\textheight]{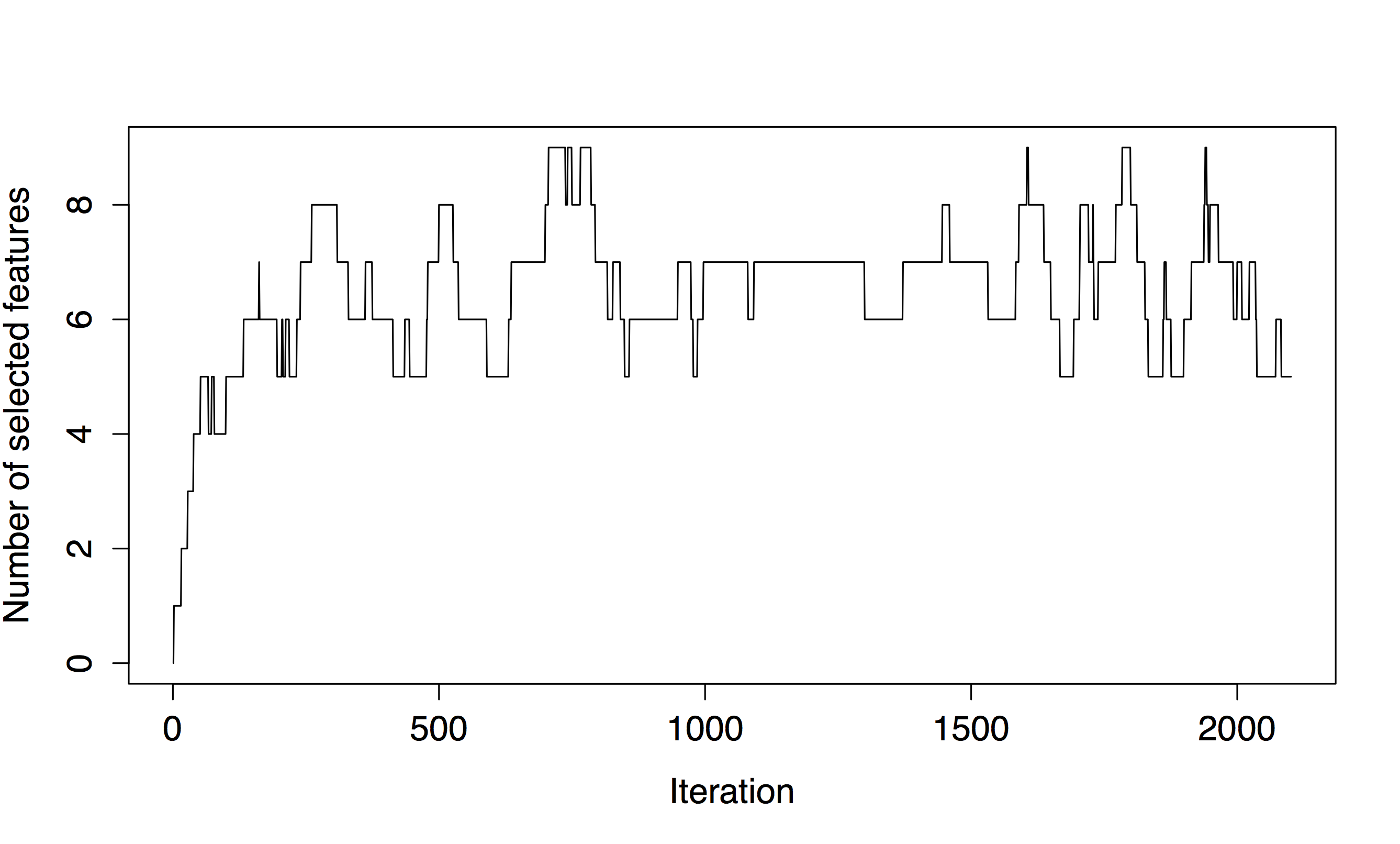}}
\vspace{.2in}
\caption{Typical realization of a simulation study for independent model and $p=10000$. Top: values of the logistic aggregation estimator on the first 100 coordinates. Bottom: number of selected features in the Metropolis-Hastings algorithm as a function of iterations, where $T_0=100$ and $T=2000$.}
\label{simfig}
\end{figure}

Figure~\ref{simfig} shows a typical behavior of the LA estimator for one particular realization on independent model and $p=10000$ with $T_0=100$ and $T=2000$. We can see from the top figure that the sparsity pattern is well recovered among the first 100 coordinates and the estimated values are close to the true value of $2$. The bottom figure displays the evolution of the Metropolis-Hastings algorithm. There exits evidence that the Metropolis-Hastings algorithm converges after only 100 iterations. \\

\begin{table}[!h]
\caption{Experimental results on the performance of logistic aggregation (LA), logistic regression (LR), $L_1$ regularization logistic regression ($L_1$-LR), and elastic net (Enet) methods; the averaged AUC, number of false positives (FP), number of false negatives (FN), computational time as well as the standard errors (in parentheses) are reported.}
\label{sim}
\begin{center}
\begin{tabular}{clccrrr}
\hline
\hline
{\bf Model} & {\bf $(n,p)$}&{\bf Method} &{\bf AUC\quad} &{\bf FP\quad} &{\bf FN\quad} &{\bf Time\text{    }}\\
\hline
\hline
Indep & $n=300$& LA & 0.941 (0.023) & 13.9 (4.8) & 0.1 (0.4) & 7.5 (0.3) \\
& $p=5000$ &  LR & 0.928 (0.036) & 0.0 (0.0) & 0.5 (0.6) & 14.6 (0.1)\\
& & $L_1$-LR & 0.921 (0.012) & 90.4 (26.2) & 0.0 (0.0) & 3.1 (0.1) \\
& & Enet & 0.916 (0.017) & 96.2 (32.6) & 0.0 (0.0) & 39.4 (1.1)\\
\cmidrule(l){2-7}
& $n=300$& LA &  0.940 (0.027) & 7.9 (3.8) & 0.1 (0.4) & 7.8 (0.3)\\
& $p=10000$ &  LR & 0.915 (0.039) & 0.0 (0.2) & 0.7 (0.7) & 29.6 (0.1)\\
& & $L_1$-LR & 0.918 (0.011) & 100.3 (26.8) & 0.0 (0.0) & 5.3 (0.1)\\
& & Enet & 0.914 (0.015) & 107.7 (36.4) & 0.0 (0.0) & 63.6 (1.5)\\
\hline
\hline
AR(1) & $n=300$& LA & 0.918 (0.030) & 13.5 (4.2) & 0.1 (0.3) & 7.5 (0.3)\\
& $p=5000$ &  LR & 0.760 (0.040) & 0.0 (0.2) & 2.7 (0.7) & 14.7 (0.2)\\
& & $L_1$-LR & 0.882 (0.018) & 88.4 (28.9) & 0.0 (0.0) & 3.2 (0.1)\\
& & Enet &  0.874 (0.025) & 85.5 (32.1) & 0.0 (0.0) & 37.7 (0.7)\\
\cmidrule(l){2-7}
& $n=300$& LA & 0.890 (0.070) & 10.8 (5.8) & 0.4 (0.8) & 8.7 (0.5) \\
& $p=10000$ &  LR & 0.742 (0.053) & 0.0 (0.2) & 2.9 (0.8) & 29.2 (0.1)\\
& & $L_1$-LR & 0.865 (0.017) & 96.7 (26.8) & 0.0 (0.0) & 5.4 (0.1)\\
& & Enet & 0.862 (0.026) & 97.2 (28.1) & 0.0 (0.0) & 61.5 (1.0)\\
\hline
\hline
AR(2) & $n=300$& LA & 0.916 (0.046) & 12.4 (4.1) & 0.2 (0.6) & 8.3 (0.6)\\
& $p=5000$ &  LR & 0.772 (0.051) & 0.0 (0.0) & 2.5 (0.8) & 18.0 (0.9)\\
& & $L_1$-LR & 0.887 (0.017) & 84.0 (30.1) & 0.0 (0.0) & 3.2 (0.1)\\
& & Enet & 0.881 (0.020) & 87.0 (32.0) & 0.0 (0.0) & 41.1 (1.6)\\
\cmidrule(l){2-7}
& $n=300$& LA & 0.911 (0.051) & 11.6 (4.9) & 0.2 (0.7) & 9.6 (0.6)\\
& $p=10000$ &  LR & 0.753 (0.050) & 0.1 (0.2) & 2.8 (0.8) & 35.0 (1.9)\\
& & $L_1$-LR & 0.878 (0.021) & 88.1 (37.7) & 0.0 (0.0) & 5.5 (0.2)\\
& & Enet & 0.871 (0.025) & 92.4 (33.8) & 0.0 (0.0) & 65.5 (2.7)\\
\hline
\hline
\end{tabular}
\end{center}
\end{table}

\section{Analysis of Real Dataset}

In this section, we apply the logistic aggregation (LA) method to a published genome-wide case-control data on Parkinson disease with 270 case individuals, 271 control individuals, and approximately 408,000 SNPs (Fung et al. 2006). The goal of genome-wide association study (GWAS) is to examine many common genetic variants in different individuals to see if any variant is associated with a trait, like major human diseases. In this investigation, we would like to identify a set of causal SNPs associated with Parkinson disease. \\

Initially we compute the $p$-value for each SNP using a single-locus logistic regression with the binary response coding as $1:=$ case and $0:=$ control, that is, testing each SNP one at a time, where an additive model on the genotype is assumed. None of the SNPs is statistically significant after the Bonferroni correction for multiple testings. The $p$-value for the ``best'' SNP rs6826751 is $2.46\times10^{-6}$, which is far from significant after Bonferroni correction. \\

To make the scale of the dataset manageable, we excluded those SNPs whose $p$-values (without correction for multiply testings) exceeded 0.01; 3857 SNPs remained in the dataset for the further analysis. Then we apply our aggregation method to this smaller dataset. In a random split, half of the data was used in the first stage for constructing estimators, while the other half was used in the second stage for aggregation. \\

\begin{figure}[!h]
\vspace{.2in}
\centering{\includegraphics[width=.675\textwidth,height=.3\textheight]{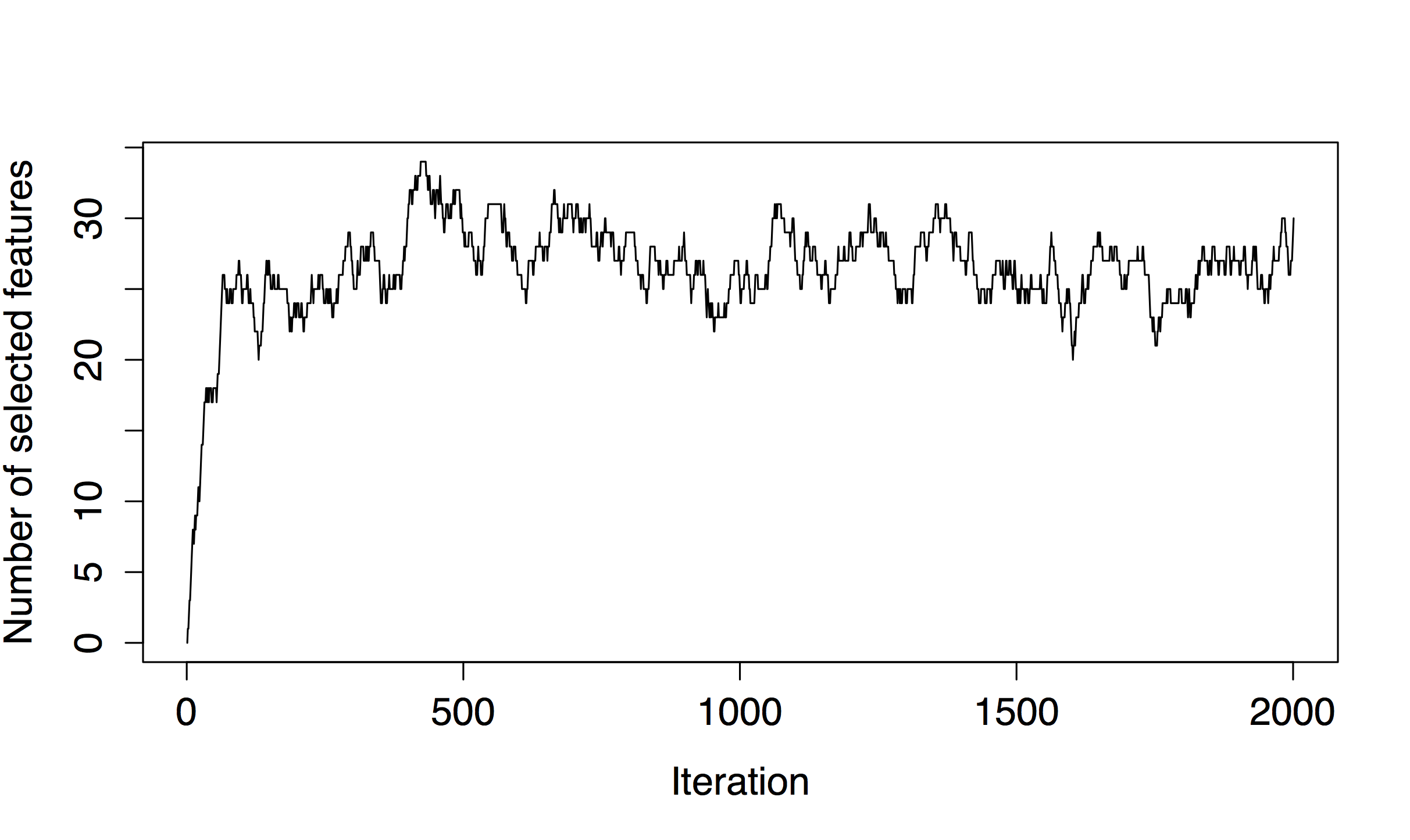}}
\vspace{.2in}
\caption{Analysis results on Parkinson disease GWAS data; number of selected features in the Metropolis-Hastings algorithm as a function of iterations, where $T_0=500$ and $T=1500$.}
\label{realfig}
\end{figure}

Figure~\ref{realfig} displays the evolution of Metropolis-Hastings algorithm where $T_0=500$ and $T=1500$. We can see that the Metropolis-Hastings algorithm converges after 500 iterations. \\

The LA estimator selected 41 SNPs. Table~\ref{snp} shows the top 10 SNPs with the largest absolute value of estimated LA coefficients, with their odds ratios (OR), marginal $p$-value using single logistic regression also reported. More detailed analysis of the biological implications of this work will left as a future study. \\

\begin{table}[!h]
\caption{Analysis results on Parkinson disease GWAS data; the top 10 SNPs with the largest absolute value of estimated LA coefficients, as well as their odds ratios (OR), marginal $p$-value using single logistic regression are reported.}
\label{snp}
\begin{center}
\begin{tabular}{rlrccr}
\hline
\hline
{\bf Chrom.} & {\bf SNP rs\#} & {\bf Position} & {\bf OR} & {\bf Marginal $p$-value} & {\bf LA Coef.}\\
\hline
10 & rs2505513CG & 42953543 & 1.77 & 1.2E-04 & 0.585 \\
22 & rs229492GA & 35885092 & 0.48 & 1.1E-04 & -0.456 \\
12 & rs7972947CA & 2040694 & 1.81 & 1.7E-04 & 0.445 \\
1 & rs7543509AG & 110764041 & 0.38 & 7.5E-05 & -0.441 \\
1 & rs7554157TC & 190220826 & 0.60 & 4.4E-05 & -0.410 \\
6 & rs719830AG & 73488690 & 0.52 & 7.6E-04 & -0.371 \\
11 & rs1912373CT & 56240441 & 0.54 & 5.5E-05 & -0.367 \\
9 & rs546171AC & 12869368 & 0.35 & 1.4E-04 & -0.362 \\   
12 & rs10879957CT & 74350612 & 1.63 & 9.3E-05 & 0.361 \\    
14 & rs7157079AG & 94397102 & 0.33 & 3.1E-04 & -0.345 \\
\hline
\hline
\end{tabular}
\end{center}
\end{table}

\section{Discussion}
In this paper, we proposed an aggregation algorithm for sparse logistic regression and demonstrated that this estimator can give comparable or better results than the $L_1$-penalized logistic regression. We show that this method could be a promising toolset in practical applications. \\

Our aggregation method is based on a sample-splitting procedure: the first subsample is set to construct the estimators and the second subsample is used to determine the component weights and aggregate these estimators. \\

One disadvantage of our method is that we sacrifice half of the data in either stage. Although multi-split procedures could be a promising solution, our approach may lose some efficiency when the sample size is relatively small. Future work should consider this in a better way. \\

\pagebreak

\end{document}